\documentclass[letterpaper, 10 pt, journal, twoside]{IEEEtran}

\IEEEoverridecommandlockouts

\usepackage{graphics} 
\usepackage{epsfig} 
\usepackage{times} 
\usepackage{amsmath} 
\usepackage{amssymb}  
\usepackage[table]{xcolor}
\usepackage[ruled,linesnumbered, noend]{algorithm2e}

\usepackage{cite}
\usepackage{comment}
\usepackage{multirow}
\usepackage{graphicx}
\usepackage{wrapfig}
\usepackage{subfigure}
\usepackage{balance}
\usepackage{wrapfig}
\usepackage{booktabs}  

\usepackage{soul, color}
\usepackage{colortbl}
\definecolor{Gray}{gray}{0.9}

\usepackage{pifont}

\newcommand{\D}[0]{DUR} 
\newcommand{\hq}[0]{HQ-Net } 
\newcommand{\gv}[0]{GViT }

\newcommand{\ve}[1]{\mathbf{#1}} 

\usepackage{pifont}
\usepackage{graphicx} 

\title{A Diffusion-based Data Generator for Training Object Recognition Models in Ultra-Range Distance}
\author{Eran Bamani, Eden Nissinman, Lisa Koenigsberg, Inbar Meir and Avishai Sintov 
\thanks{E. Bamani, E. Nissinman, L. Koenigsberg, I. Meir  and  A. Sintov are with the School of Mechanical Engineering, Tel-Aviv University, Israel. 
Corresponding Author: sintov1@tauex.tau.ac.il.}
}

\date{November 2023}

\begin{document}

\maketitle

\begin{abstract}

Object recognition, commonly performed by a camera, is a fundamental requirement for robots to complete complex tasks. Some tasks require recognizing objects far from the robot's camera. A challenging example is Ultra-Range Gesture Recognition (URGR) in human-robot interaction where the user exhibits directive gestures at a distance of up to 25~m from the robot. However, training a model to recognize hardly visible objects located in ultra-range requires an exhaustive collection of a significant amount of labeled samples. The generation of synthetic training datasets is a recent solution to the lack of real-world data, while unable to properly replicate the realistic visual characteristics of distant objects in images. In this letter, we propose the Diffusion in Ultra-Range (DUR) framework based on a Diffusion model to generate labeled images of distant objects in various scenes. The DUR generator receives a desired distance and class (e.g., gesture) and outputs a corresponding synthetic image. We apply DUR to train a URGR model with directive gestures in which fine details of the gesturing hand are challenging to distinguish. DUR is compared to other types of generative models showcasing superiority both in fidelity and in recognition success rate when training a URGR model. More importantly, training a DUR model on a limited amount of real data and then using it to generate synthetic data for training a URGR model outperforms directly training the URGR model on real data. The synthetic-based URGR model is also demonstrated in gesture-based direction of a ground robot.


\end{abstract}

\section{Introduction}

Visual object recognition is essential for robots to interact effectively with their environment. Image resolution degradation caused by the increased distance between the robot and the object of interest can significantly hinder recognition performance due to the loss of discriminative visual features \cite{serre2007robust}. Furthermore, cluttered backgrounds with numerous objects can pose a significant challenge for accurate object identification algorithms \cite{david2005object}. Recognition in long distances is imperative in various other applications such as surveillance \cite{shirke2019biometric}, and the detection of drones \cite{Zitar2023}, drowning scenarios \cite{lu2004vision} and ships \cite{yang2018automatic}. Particularly in robotics and Human-Robot Interaction (HRI), gesture recognition from a large distance is required for a user to efficiently convey natural directives to a robot. 
\begin{figure}
    \centering
    \includegraphics[width=\linewidth]{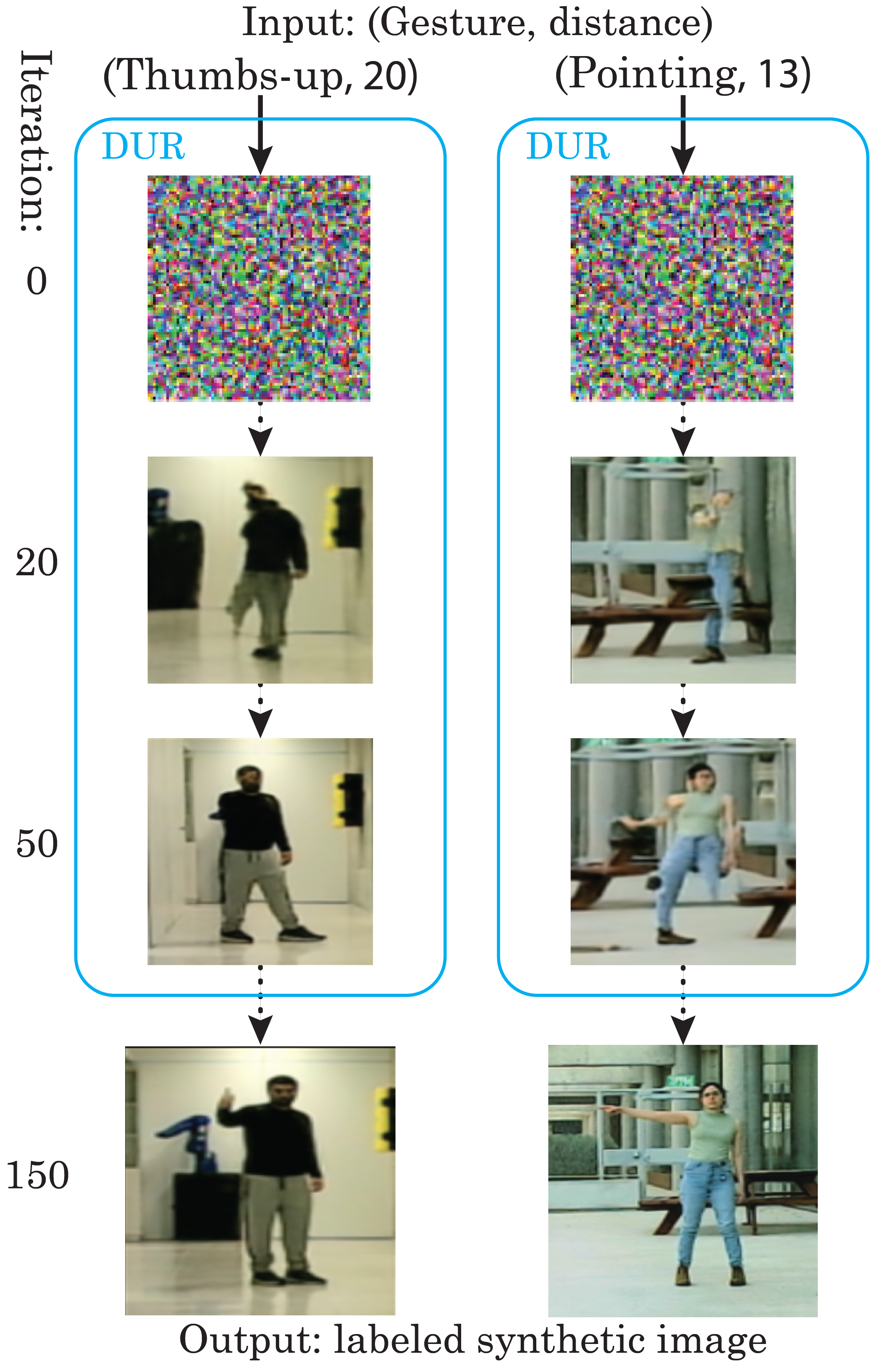}
    \vspace{-0.6cm}
    \caption{Two examples of generating synthetic images in DUR with gesture and distance conditions. Across 150 iterations, Gaussian noise is transformed into a comprehensive high-fidelity image according to the desired labels.}
    \label{fig:itr_gen}
    \vspace{-0.6cm}
\end{figure}

While state-of-the-art in gesture recognition provides an effective distance of up to seven meters and mostly indoor \cite{zhou2021long, nickel2007visual, wachs2011vision, xia2019vision, yi2018long}, recent work by the authors have addressed the Ultra-Range Gesture Recognition (URGR) problem demonstrating an effective distance of up to 25 meters in various indoor and outdoor environments \cite{bamani2024ultra}. A central challenge in URGR, and in other ultra-range object recognition problems, is the acquisition of data across varying environmental conditions and distances, which is instrumental for feasible model training. Constraints on data acquisition often originate from limited time and human resources, insufficient equipment, or low variability in environmental contexts.
\begin{figure*}
    \centering
    \includegraphics[width=\linewidth]{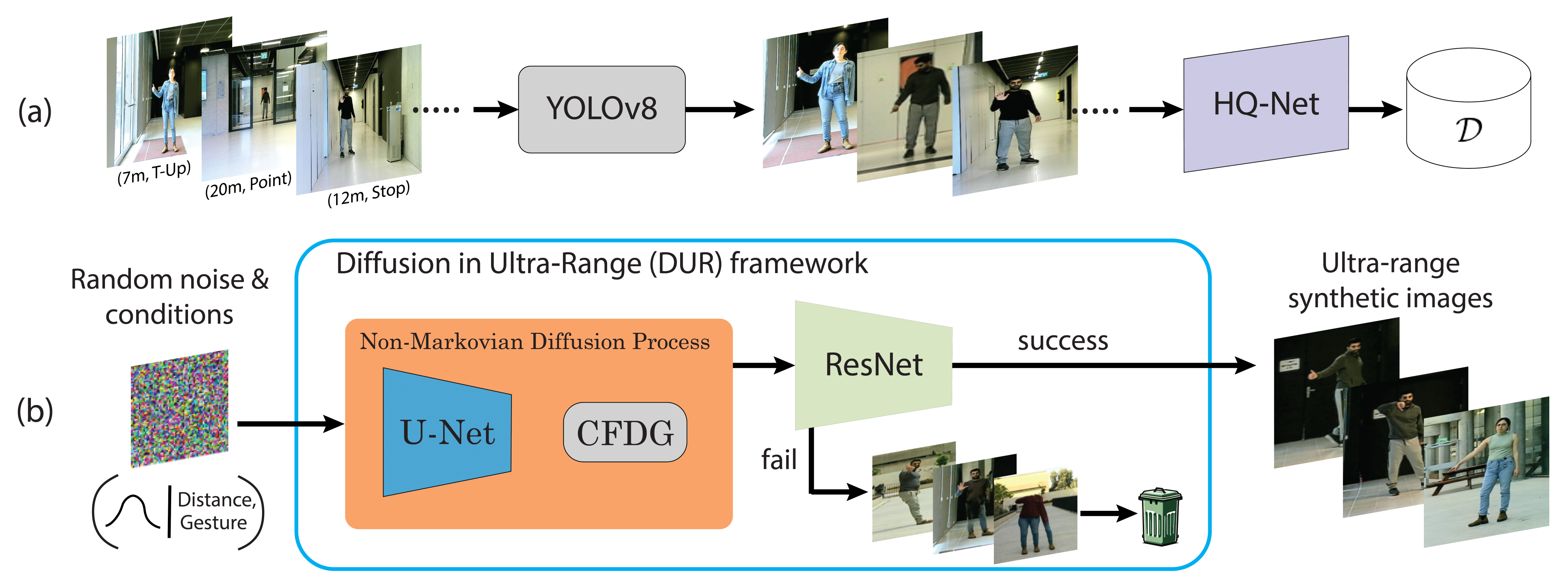}
    \vspace{-0.7cm}
    \caption{Illustration of the Diffusion in Ultra-Range (DUR) framework. (a) Dataset $\mathcal{D}$ for training DUR is acquired by collecting labeled RGB images, cropping the user with YOLOv8, and improving image quality with HQ-Net. (b) The generation of conditional synthetic samples starts with Gaussian noise along with distance and gesture conditions. A non-Markovian diffusion process denoises the noisy image and the ResNet filter removes failed images.}
    \label{fig:scheme}
    \vspace{-0.5cm}
\end{figure*}

To effectively address these data acquisition challenges, a potential solution emerged in the form of synthetic data generation \cite{lu2024}. The approach is based on the simulation of data that closely resembles real-world samples. Some work is based on a graphical simulator (e.g., Unity and Blender) \cite{molina2014natural, Tsai8252982, de2020vision,ninos2021synthetic} which, however, cannot provide sufficiently realistic images. Over the recent years, numerous data-based techniques have arisen \cite{zhou2022lightweight} including the prominent Generative Adversarial Network (GAN) \cite{goodfellow2014generative}. GAN has revolutionized the landscape of artificial intelligence by introducing an intricate interplay between a generative model and its discriminative counterpart. Consequently, advanced versions of GAN were proposed such as Deep Convolutional GAN (DCGAN) \cite{radford2015unsupervised} and Progressive GAN \cite{karras2017progressive}. Conditional GAN (CGAN) \cite{mirza2014conditional} is a relevant model that can incorporate conditional variables, such as class labels, over the generative process and thereby synthesizing specifically targeted data with predefined attributes. In the context of object recognition, GAN variants have been proposed to improve recognition in, for instance, low-quality images \cite{Prakash2021} and strong lighting conditions \cite{minciullo2021db}. In human gestures, several works used GAN to augment the training data and improve recognition rates \cite{tang2018gesturegan,fang2019gesture,liu2019improved,shen2021imaginative, waheed2020covidgan}. However, the approaches focus on close-range recognition where the exerted gesture features are observable in high quality and quite easily distinguishable.

While demonstrated efficacy in various applications, the training of GAN requires substantial computational resources, extensive datasets, and long training time  \cite{borji2019pros}. In addition, the training of GANs may exhibit sensitivity to hyperparameter selection, suffer from mode collapse, and demonstrate reduced effectiveness in generating complex visual data, particularly for objects viewed relatively far within the image \cite{hellermann2021leveraging}. To address these challenges, diffusion models have recently been introduced and emerged as a powerful tool for generative tasks \cite{sohl2015deep}. Unlike GAN and other generative methods, diffusion models are able to generate diverse high-quality data with fewer stability issues \cite{Croitoru2023}. In addition, they do not have a low-dimensional latent space while modeling it as a random Gaussian distribution to maintain the original image size. Diffusion models operate by iteratively adding noise to data samples and then denoising them, thereby capturing the underlying data distribution \cite{song2019generative}. By simulating diffusion processes, intricate patterns within the data are learned. 

Due to their versatility and effectiveness, diffusion models are used in various domains, including image denoising \cite{ho2020denoising}, super-resolution \cite{yue2023resshift}, and inpainting \cite{lugmayr2022repaint}. Few attempts have used diffusion models in object detection applications \cite{Wang2024}. For instance, it was used to augment weed identification \cite{Chen2024}. In a different work, noisy bounding boxes on the image are denoised to boxes on objects of interest \cite{Chen2023}. Nevertheless, these applications do not consider low-resolution images with specific importance to the detection of small features. To the best of the authors' knowledge, work is absent with diffusion models for conditional and ultra-range image generation relevant to object and gesture recognition. The inherent problem of generating synthetic human gesture data across diverse environments is further challenged when solely focusing on the distance variation from the viewpoint of the camera.

In this work, we address the generation of conditional synthetic images with desired objects of interest located in the ultra range. Within this effort, we propose a novel framework termed \textit{Diffusion in Ultra-Range} (DUR). DUR is a generator of conditional synthetic images that contain desired objects viewed in ultra-range from the camera (Figure \ref{fig:itr_gen}). Hence, the object in the synthetic image is seen far from the viewpoint of the camera and in low quality, making it difficult to identify fine details, as in the real-world case. DUR, illustrated in Figure \ref{fig:scheme}, is based on a Diffusion model and specifically on non-Markovian Diffusion processes for enhanced performance \cite{song2020denoising}. With the DUR generator, one can generate large-scale synthetic datasets for training ultra-range object recognition models. As a test case, we explore the efficacy of synthetic images generated by DUR, compared to real data, for training a URGR model. In the URGR problem, the target object for recognition is small and exhibits low-resolution characteristics, such as blurry hand features, within a visually complex environment. This specific application presents a stringent evaluation benchmark for the DUR generator's capability to synthesize realistic images suitable for training a robust object recognition model. The key contributions of this work are:
\begin{itemize}
    \item We propose the novel DUR framework for generating synthetic and labeled imagery datasets with objects of interest situated in an ultra-range distance from the viewpoint of the camera. 
    \item As a rigorous test case, we show the ability of a DUR generator to synthesize high-fidelity imagery data of individuals exhibiting directive gestures. The generator offers control over two key conditions, allowing practitioners to manipulate the model to produce images spanning distances from 4 to 25 meters, across different gestures, environments, and human subjects.
    \item A synthetic dataset of ultra-range gestures is used to train a URGR model, exhibiting high performance compared to the same model trained with real data.
    \item The trained URGR model is demonstrated in directing a ground robot.
    \item The trained model and labeled dataset will be made available open-source for the benefit of the community\footnote{The availability of the dataset and model will be provided upon acceptance for publication, with images modified to safeguard the privacy of participants.}.
\end{itemize}

\label{sec:introduction}



\section{Methods}

\subsection{Problem Formulation}
\label{sec:problem_def}

For a robot to comprehend its environment, it is essential to train models capable of recognizing objects across diverse and challenging environments, and at varying distances in particular. Based on the URGR problem originally presented by the authors \cite{bamani2024ultra}, we define the general ultra-range object recognition problem. Given an RGB image $\ve{x}_i$ taken while an object is in a distance of $d\leq25$ meters, it is required to classify the image based on the exhibited object out of $m$ possible classes $\{\mathcal{O}_1,\ldots,\mathcal{O}_m\}$.


A significant challenge lies in generating a sufficient amount of reliable data that can effectively generalize across different environments and distances while accommodating the various object classes. 
To acquire accurate recognition, a substantial amount of heterogeneous data is required, covering different objects, poses, environments, camera angles and distances. However, acquiring sufficient data may not be feasible due to limited resources and time. Therefore, the objective of this work is to train a synthetic image generator capable of generating diverse images in varied environments and distances. In practice, the generator $\Theta(\mathcal{O}_i,d)$ will output a synthetic image exhibiting a desired object class $\mathcal{O}_i$ in distance $d$. The ability to manipulate both the object classes and the distance from which they are captured will provide the capacity for generating a large amount of diverse training data. In this work, we specifically evaluate a generator for the URGR problem where the distinction of finger features during gesture is challenging.

\subsection{Data Collection}
\label{sec:data_collection}

To train the generative gesture model, we collect a labeled dataset $\mathcal{D}=\{(\ve{x}_i,o_i, d_i)\}_{i=1}^N$. Each image $\ve{x}_i$ in $\mathcal{D}$ is labeled with the gesture class index $i\in\{1,\ldots,m\}$ and distance $d_i\in[4,25]$. Images are taken using a simple web camera with several users in various environments. Particularly, the dataset was acquired from diverse indoor and outdoor settings, with users moving freely within the designated range while performing one of several possible gestures. To have an image focused on the user, each image in the dataset goes through pre-processing by detecting the user within it using YOLOv8 and cropping out the background. When considering a general object recognition problem, manual segmentation of the objects of interest would replace the YOLO step. Since the cropped images are of low resolution, we augment their quality with HQ-Net \cite{bamani2024ultra}. \hq is a Super Resolution (SR) method that uses a set of filters, self-attention mechanisms, and convolutional layers to improve the compromised image fidelity of a distant object in the image. The collection process is illustrated in Figure \ref{fig:scheme}a.

\subsection{Diffusion models}
\label{sec:background}

The diffusion model is a fundamental technique for the generation of synthetic images. An intricate interplay of diffusion processes contributes to the creation of visually compelling and high-fidelity images. Unlike GAN, the diffusion model excels in generating diverse and high-quality data. We briefly discuss the fundamentals of the Denoising Diffusion Probabilistic Model (DDPM) while an in-depth presentation can be reviewed in \cite{ho2020denoising}. 

The DDPM captures information spread over time and space by gradually introducing and then reversing noise in the data. Hence, DDPM encompasses two processes where, in the first, a training image is destroyed by iteratively adding noise over time, making it progressively less recognizable until it transforms into pure noise. In the second process, the noising process is reversed. The reverse process converts the complex distribution back to a simple distribution using inverse transformations. Hence, the generation of a synthetic image can be done by randomly sampling noise and passing it through the reverse process.

The forward process can be considered as a Markov chain. An image is sampled from the real data distribution $\ve{x}_0 \sim q(\ve{x})$ and corrupted by gradually adding Gaussian noise for $T$ steps yielding a sequence of noisy samples $\mathbf{x}_1, \ldots, \mathbf{x}_T$. Let $\{\beta_t \in (0, 1)\}_{t=1}^T$ be a sequence of learned or fixed variance schedules regulating the noise level, the forward process is given by
\begin{equation}
    p(\ve{x}_{1:T} | \ve{x}_{0}) = \prod^T_{t=1}p(\ve{x}_t | \ve{x}_{t-1})
\end{equation}
where
\begin{equation}
    \label{eq:ptt}
    p(\ve{x}_t | \ve{x}_{t-1}) = \mathcal{N}(\ve{x}_t;~ \ve{x}_{t-1}\sqrt{1 - \beta_t}, \beta_t\ve{I}).
\end{equation}

In the reversed process, a noise sample $\ve{x}_T\sim\mathcal{N}(0,\ve{I})$ is used to recreate an image from the original distribution. However, an inverse probability map $p(\ve{x}_{t-1}|\ve{x}_t)$ is not available. Hence, a model $p_\theta$ is learned to approximate the conditional probabilities as
\begin{equation}
    p_\theta(\ve{x}_{t-1} | \ve{x}_t) = \mathcal{N}(\ve{x}_{t-1};~ \mu_\theta(\ve{x}_t,t),\Sigma_\theta(\ve{x}_t,t))
\end{equation}
where $\mu_\theta$ and $\Sigma_\theta$ are time-dependent mean and variance parameters of the Gaussian transition from $\ve{x}_t$ to $\ve{x}_{t-1}$. While the variance is usually chosen to be $\Sigma_\theta(\ve{x}_t,t)=\beta_t\ve{I}$, the diffusion posterior mean $\mu_\theta(\ve{x}_t,t)$ is implemented using a neural network model which is required to have equal dimensionality at both input and output. 

While Markovian diffusion models provide high generative capabilities, sampling a synthetic image tends to be significantly slow in the reverse process over thousands of iteration steps $T$. Hence, the Denoising Diffusion Implicit Model (DDIM) model was proposed where a non-Markovian process is considered \cite{song2020denoising}. In such case, the recovery iteration step in \eqref{eq:ptt} is reformulated to
\begin{equation}
    \label{eq:ptt0}
    p_\theta(\ve{x}_{t-1} | \ve{x}_t, \ve{x}_0) = \mathcal{N}(\ve{x}_{t-1};~ \tilde{\mu}_\theta(\ve{x}_t,\ve{x}_0),\beta_t\ve{I})
\end{equation}
where $\tilde{\mu}_\theta(\ve{x}_t,\ve{x}_0)$ is learned using a neural network. Such a formulation has been shown to provide high-quality synthetic images in a faster time and provides deterministic synthesis. 

The neural networks of both DDPM and DDIM are commonly based on the U-Net architecture \cite{ronneberger2015u}. U-Net is trained to reverse the Markov transitions in the forward process to maximize the likelihood of the training data. By leveraging the U-Net architecture in the reverse process, the diffusion model can effectively reconstruct high-fidelity images from the noise sequences generated during the forward process. This integration highlights the synergy between advanced neural network architectures and probabilistic models, leading to robust and reliable image generation techniques.

\subsection{DUR framework}

To synthesize labeled data for training a gesture recognition model in ultra-range, we propose the Diffusion in Ultra-Range (DUR) framework illustrated in Figure \ref{fig:scheme}b. The framework uses the DDIM to generate conditional synthetic images and a ResNet-based model \cite{he2015deep} as an image quality control module. To learn the transition component of DDIM in \eqref{eq:ptt0}, the U-Net architecture was used. The unique architecture of U-Net is characterized by a contracting path followed by an expansive path. During the expansive path, high-resolution features are gradually restored by combining information from multiple scales, enabling the network to recover fine details lost during the diffusion process. Furthermore, skip connections enable the direct flow of information from the contracting path to the expansive path aiding in preserving spatial information and ensuring precise reconstruction. This mechanism is particularly beneficial in the reverse process of the diffusion model, where the goal is to faithfully reconstruct an image from the noise-corrupted version. 

The U-Net in DUR incorporates temporal encoding to generate accurate outputs corresponding to specific time points in the diffusion process. This temporal information is fed into the U-Net as a condition through sinusoidal positional embedding. To constrain the model to generate images of desired classes and distances, two additional conditions, e.g., gesture and distance, are added to the time embedding. Hence, the input to the U-Net is an embedding of the data
\begin{equation}
    u_{emb} = \ve{x}_{emb} + o_{emb} + d_{emb}    
\end{equation}
where $o_{emb}$ and $d_{emb}$ are the vector embeddings of the class and distance conditions, respectively. These are added to $\ve{x}_{emb}$, a down-sampled embedding of the noise image $\ve{x}_T\sim\mathcal{N}(0,\ve{I})$. The sampling process of images is required to be of low temperature. In other words, the probability of sampling an image not in the desired class will be low in a rather deterministic process. Hence, we employ the Classifier-Free Diffusion Guidance (CFDG) \cite{ho2022classifier}. The guidance enhances learning and improves sample quality.

\begin{figure}
    \centering
    \includegraphics[width=\linewidth]{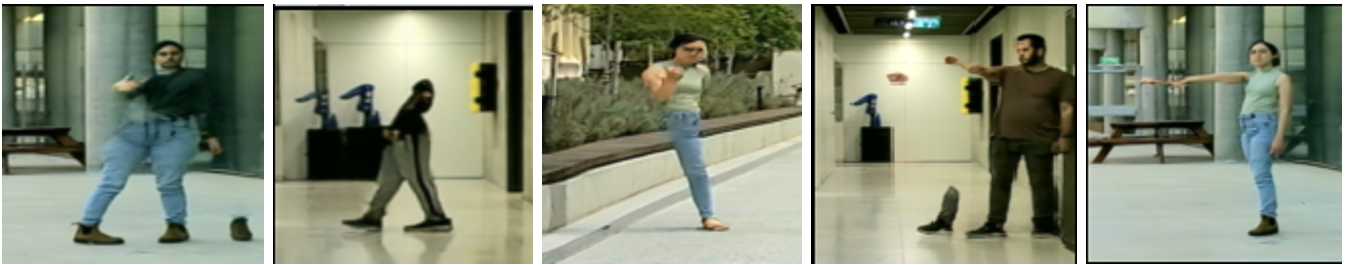} 
    \caption{Examples of failed synthetic images generated by DUR.}
    \label{fig:fail}
    \vspace{-0.5cm}
\end{figure}

With this diffusion formulation, one can generate high-fidelity and realistic images while enforcing a desired gesture and user distance from the camera, as seen in Figure \ref{fig:itr_gen}. However, preliminary observations of the outputted images from the diffusion model have revealed that some portions are partly unrealistic or distorted. For instance, the user may be generated without a limb or smeared. Examples of such images can be seen in Figure \ref{fig:fail}. Hence, the DUR framework includes an image quality control model used to filter low-quality images. A systematic evaluation of image quality is conducted to optimize the results of the generative process. After training the DDIM, $K$ synthetic images are generated. Then, they are manually labeled as successful or failed ones. An image is labeled as failed if the structural integrity of the user in the image is distorted or missing. With the labeled images, we train a 32-layer ResNet with embedding to recognize successful and failed images. The trained ResNet is then a filter for unfit images generated by the DDIM. Hence, it is used to mitigate low-quality data, such as images depicting only partial body parts
, diminished image quality, smoothing effects and other complexities. 

\label{sec:method}

\section{Model Evaluation}

In this section, we evaluate the proposed DUR framework for generating conditional synthetic data and using them for training URGR models. Our analysis focuses on six distinct gesture classes that can be used to direct a robot: pointing, thumbs-up, thumbs-down, beckoning, stop and null, observed across a range extending from 4 to 25 meters. 
Null is the lack of an exhibited gesture where the user can perform any other task. We next analyze the quality of the generated images in DUR followed by assessing their usage in training gesture recognition models. All computations were performed on a Linux Ubuntu (18.04 LTS) machine with an Intel Xeon Gold 6230R CPU (consisting of 20 cores operating at 2.1GHz) and four NVIDIA GeForce RTX 2080TI GPUs, each endowed with 11GB of RAM.



\subsection{Dataset}

Dataset $\mathcal{D}$ of real images with gesture and distance labels was obtained using eight users of diverse ages and genders. The users alternately executed different gestures out of the six in various in-door and out-door environments. The exhibited gesture can be performed with either the right or left arm. In addition, the data was collected in a nearly uniform distribution along the $d\in[4,25]$ meters distance range. Image acquisition involved a web camera producing images sized at $480 \times 640$ pixels. The collection yielded $N=175,000$ labeled samples with approximately 29,105 samples per gesture. Furthermore, a separate test set comprising 5,821 labeled images was gathered from 16 users not included in the training process and in different environments.




\subsection{Synthetic image quality}
\label{sec:quaility}

Using dataset $\mathcal{D}$, the DUR framework is trained. We first analyze the quality and realistic characteristics of the synthetic images from DUR compared to synthetic images from other generative models. Our comparison includes CGAN,  Auxiliary Classifier GAN \cite{odena2017conditional}, DCGAN and Progressive GAN. While CGAN and ACGAN can handle conditioning of the generated images, DCGAN and Progressive GAN were modified to align with the specific input conditions of gestures and distances. In brief, both the generators and discriminators of these two GAN's were added an input of the gesture class and distance. We also compare to DUR without filtering out failed images with the ResNet model. The ResNet was trained with $K=16,800$ manually labeled images.

\begin{figure*}
    \centering
    \includegraphics[width=\linewidth]{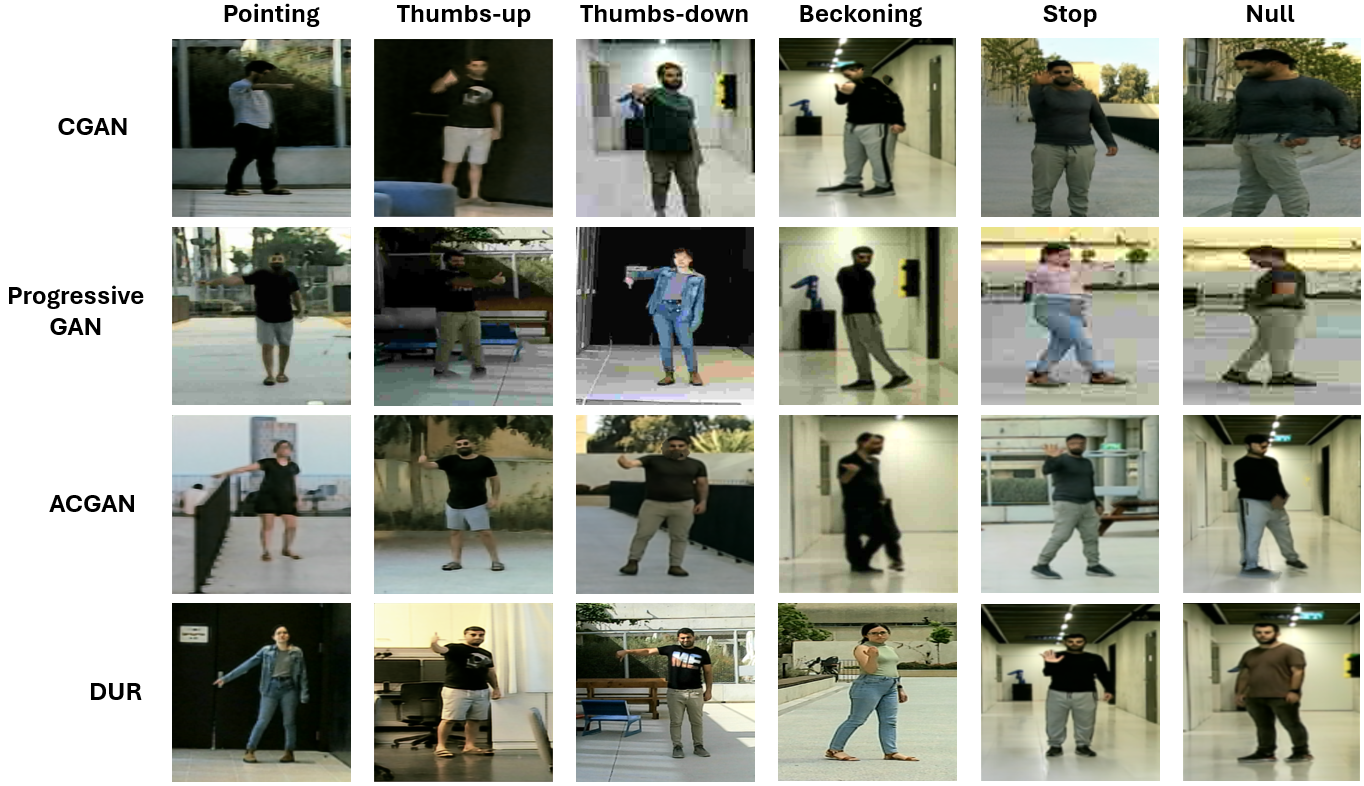}
    \vspace{-0.7cm}
    \caption{Visual comparison of synthetic image examples across several generative models, including DUR, and for the trained gesture classes.}
    \label{fig:syn_compare}
\end{figure*}

All GAN models were trained with the same data in $\mathcal{D}$. Then, each model, including DUR, generated a set of synthetic images with an amount equal to the size of the test set. We evaluate the quality of the synthetic images, some with respect to the test data, using three metrics: Inception Score (IS) \cite{salimans2016improved}, Fréchet Inception Distance (FID) \cite{heusel2017gans} and Structural Similarity Index (SSIM) \cite{wang2004image}. IS evaluates the quality of synthetic images and considers both the variety and distinctiveness of the generated images, with a higher score indicating more realistic and diverse outputs. On the other hand, FID is a common pixel-level metric of domain adaptation that compares the statistical features of real and generated images without requiring paired image datasets. A lower FID score indicates that the generated images better capture the complexity and realism of the real data. Similarly, SSIM does not require paired images and evaluates how similar two images are in terms of luminance, contrast, and structure, aiming to reflect how humans perceive the difference. Hence, a score closer to 1 indicates high similarity. Table \ref{tb:Different_Metrics} summarizes the comparative results for all generative models and metrics. DUR is shown to be significantly superior compared to all GAN models across all metrics. In addition, ResNet is validated to considerably improve the quality of synthetic images in the DUR framework. Figure \ref{fig:syn_compare} provides a visual comparison of synthetic image examples generated by the above generative models. The examples show that DUR provides the most realistic and physically accurate samples.

\begin{table}
\centering
\caption{Comparative quality evaluation of generative models}
\label{tb:Different_Metrics}
\begin{tabular}{lcccc}\toprule
        Models  & & IS & FID & SSIM \\\midrule
        CGAN & \cite{mirza2014conditional}              & 6.41 & 28.46 & 0.671 \\
        ACGAN & \cite{odena2017conditional}             & 7.13 & 25.91 & 0.711 \\   
        DCGAN & \cite{radford2015unsupervised}          & 5.92 & 31.28 & 0.543 \\
        Progressive GAN & \cite{karras2017progressive}  & 7.86 & 26.26 & 0.698 \\
        \D~ w/o ResNet &                                & 8.07 & 25.31 & 0.745 \\
        \rowcolor{lightgray}
        \D &                                            & 9.47 & 20.05 & 0.907 \\
\bottomrule
\end{tabular}
\end{table}




\subsection{Usage for Gesture Recognition}

The effectiveness of synthetic images is next evaluated for training a gesture recognition model. We base our analysis on \gv previously proposed by the authors for URGR \cite{bamani2024ultra}. \gv combines the merits of two different models: Graph Convolutional Network (GCN) \cite{ullah2022graph} and Vision Transformer (ViT) \cite{dosovitskiy2021image}. This allows effective analysis of both the visual features of gestures captured by a camera and the relationships between different parts of the hand or body involved in the gesture, even at extended distances. Since a user is observed from a far distance, the captured image is of low resolution. Hence, the image goes through the super-resolution \hq model for quality enhancement before recognition with GViT. 

\begin{figure*}
    \centering
    \begin{tabular}{cccccc}
        \includegraphics[height=0.15\linewidth]{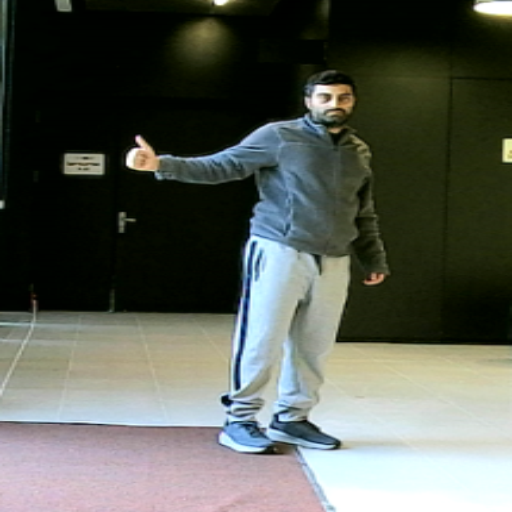} \hspace{-8px} & 
        \includegraphics[height=0.15\linewidth]{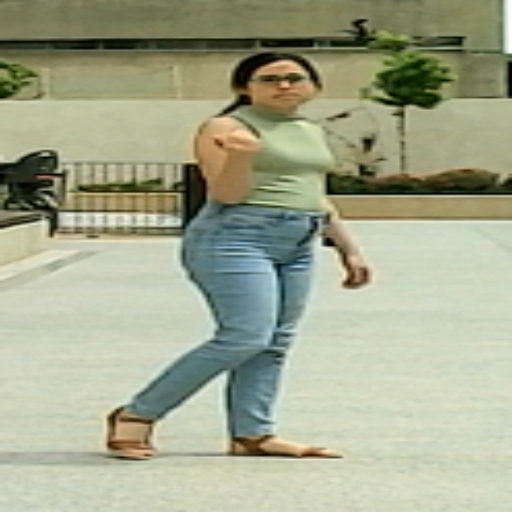} \hspace{-8px} & 
        \includegraphics[height=0.15\linewidth]{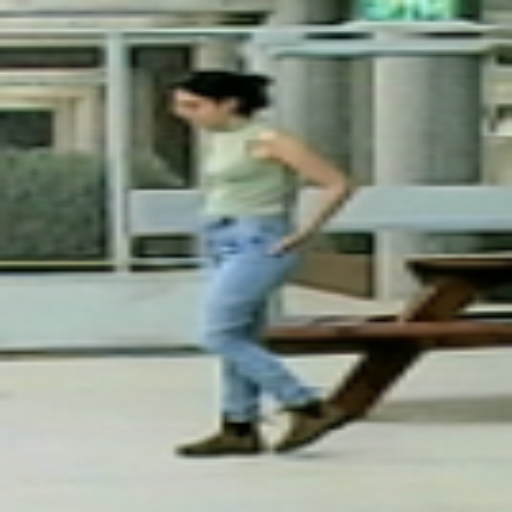} \hspace{-8px} &
        \includegraphics[height=0.15\linewidth]{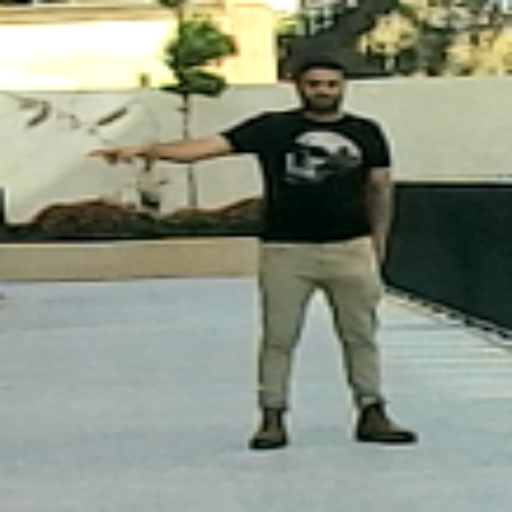} \hspace{-8px} &
        \includegraphics[height=0.15\linewidth]{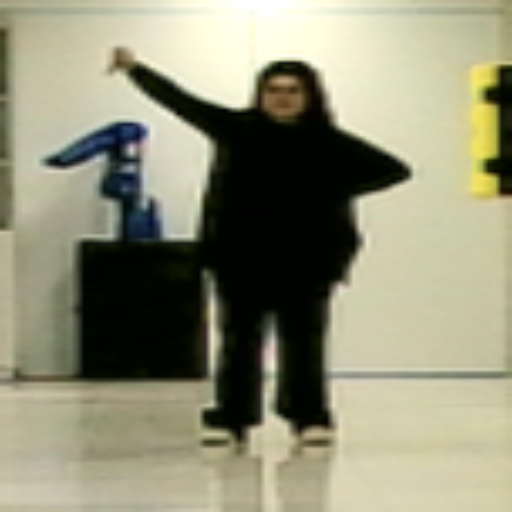} \hspace{-8px}  &  \includegraphics[height=0.15\linewidth]{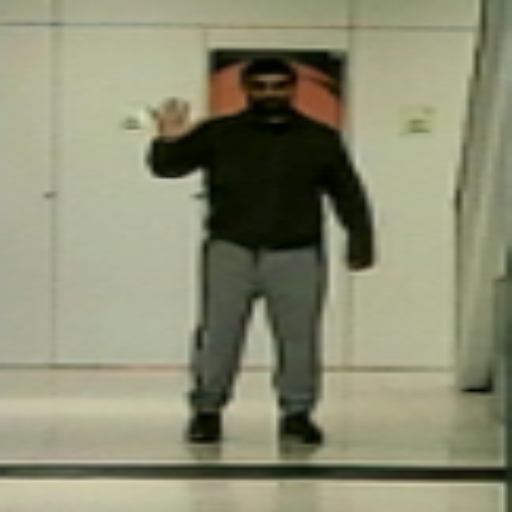} \\
    \end{tabular}
    \caption{Examples of correct gesture recognition with GViT trained on synthetic data from DUR (left to right): thumbs-up gesture from 10 meters distance with model certainty of 95.6\%; beckoning gesture from 12 meters distance with model certainty of 95.1\%; null gesture from 19 meters distance with model certainty of 93.5\%; pointing gesture from 15 meters distance with model certainty of 94.7\%;  thumbs-down gesture from 20 meters distance with model certainty of 92.8\%; and stop gesture from 17 meters distance with model certainty of 93.9\%.}
    \label{fig:dis_exm}
    \vspace{-0.5cm}
\end{figure*}

First, \gv was trained with 175,000 real images from $\mathcal{D}$ as done in \cite{bamani2024ultra}. In addition, for DUR and for each of the GAN models discussed in Section \ref{sec:quaility}, we generated a dataset of 175,000 labeled synthetic images. The datasets have a uniform distribution of sampled gestures and distances. 
A \gv model was trained with each dataset and evaluated on a real test set of 10,109 labeled images originating from the work in \cite{bamani2024ultra} and independent of any training data. Table \ref{tb:Different_IQ} presents the recognition success rate for the GViT models trained with the different sources of labeled data. The success rate for each model is averaged over five training trials. The results show that all generative models provide better recognition accuracy compared to a model with real data. This highlights the merit of investing real collected data in a generative model, compared to directly using the data for training the URGR model. Among the generative models, DUR provides a superior recognition success rate by a significant margin. Figure \ref{fig:dis_exm} demonstrates several correct gesture recognitions with GViT trained on synthetic data from DUR.
\begin{table}
\centering
\caption{Gesture recognition success rate using \gv trained with synthetic data from different generative models}
\label{tb:Different_IQ}
\begin{tabular}{lcc}\toprule
        Training data source           & & Success rate (\%)  \\\midrule
        Directly w/ real data       & \cite{bamani2024ultra}         & 58.8 $\pm$ 8.30\\
        CGAN            & \cite{mirza2014conditional}    & 78.3 $\pm$ 4.82 \\
        ACGAN           & \cite{odena2017conditional}    & 83.8 $\pm$ 5.03 \\
        DCGAN           & \cite{radford2015unsupervised} & 64.7 $\pm$ 5.71 \\
        Progressive GAN & \cite{karras2017progressive}   & 81.4 $\pm$ 3.48 \\
        \rowcolor{lightgray}
        \D              &                                & 95.8 $\pm$ 4.01 \\
\bottomrule
\end{tabular}
\vspace{-0.5cm}
\end{table}

\begin{figure}
    \centering
    \includegraphics[width=\linewidth]{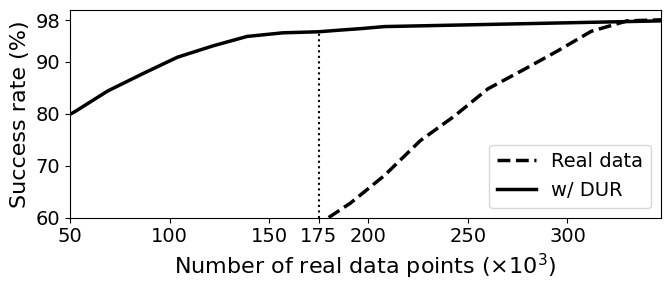}
    \vspace{-0.6cm}
    \caption{Gesture recognition success rate for training GViT with regard to the amount of real data points used to directly train GViT, or to train DUR and then generate 220,000 synthetic data for training GViT.}
    \label{fig:compare_accuracy_GVIT}
    \vspace{-0.5cm}
\end{figure}

\begin{figure}
    \centering
    \includegraphics[width=0.95\linewidth]{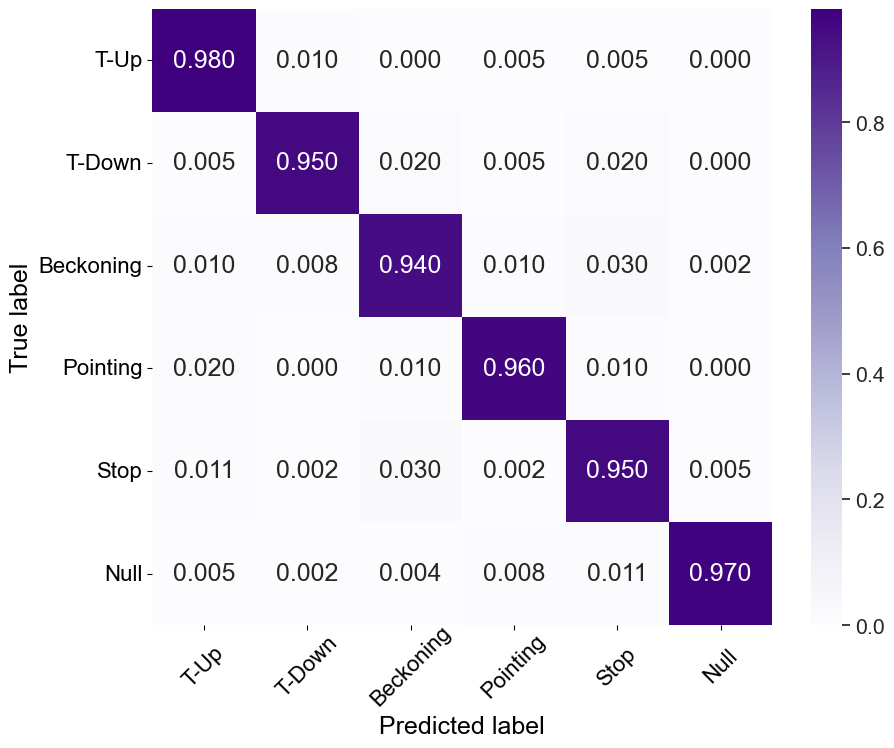} 
    \vspace{-0.4cm}
    \caption{Recognition confusion matrix for \gv trained on synthetic data from DUR and evaluated on real test data.}
    \label{fig:confmat}
\end{figure}

We further evaluate the merit of investing real collected data $\mathcal{D}$ directly to train GViT, or indirectly by first training DUR with $\mathcal{D}$ and then training GViT with synthetic data. Figure \ref{fig:compare_accuracy_GVIT} presents the success rate of gesture recognition over the test data with regard to the amount of real data used to train either DUR or directly GViT. When DUR is trained with some amount of real data, 220,000 synthetic images are then generated and used to train a GViT model. The results emphasize a significant improvement attributable to the use of a generative model. By using 175,000 real images to train DUR, the success rate reaches 95.8\%. To reach the same rate with direct GViT training, 312,000 real labeled images are required. On the other hand, using the 175,000 real images to directly train GViT yields a poor success rate of 58.8\%. 
Hence, it is significantly more advantageous to use the collected data to train DUR than directly using it to train GViT. This result can be explained by the high quality and rich images generated by DUR compared to the real data which is rather limited. DUR is able to capture the gesture image distribution with a limited number of real samples and, thus, generalize to provide representative synthetic samples from the whole distribution. Figure \ref{fig:confmat} presents the confusion matrix for the GViT trained with synthetic data from DUR. The results show high success rates for all gestures.

\begin{figure}
    \centering
    \includegraphics[width=\linewidth]{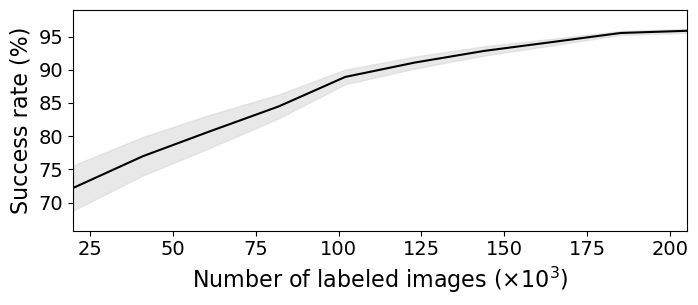} 
    \vspace{-0.7cm}
    \caption{Mean success rate for Gesture recognition for \gv with regard to the amount of synthetic training samples generated by DUR.}
    \label{fig:accuracy_GVIT_data}
    \vspace{-0.5cm}
\end{figure}

In an opposite perspective, we are interested in the implication of generating more synthetic samples to train GViT, with an already trained DUR. Hence, we use a DUR trained with 175,000 real images to generate synthetic samples. Figure \ref{fig:accuracy_GVIT_data} presents the mean success rate for gesture recognition in GViT trained with an increasing number of synthetic data. The mean success rate is evaluated over five training trials with newly generated synthetic data in each trial. The results clearly show performance improvement with an increase in synthetic samples, emphasizing the benefit of using DUR.

Next, an ablation study is provided. We compare the recognition success rate of GViT trained with synthetic data from different configurations of DUR: training DUR without passing the cropped images through HQ-Net, training DUR without CFDG, and DUR without filtering out failed images using ResNet. In addition, we also evaluate ill-trained GViT on failed synthetic test images. That is, 10,176 failed images were added to the training of the GViT. Then, it was evaluated on test images composed of 16,464 failed synthetic images. The failed images are the ones acquired for training the ResNet (Figure \ref{fig:fail}). Table \ref{tb:Gvit} presents the results for the comparison along with the one for the full DUR model. The results show that DUR, in all cases, is able to acquire relatively successful training data for GViT. Nevertheless, the SR component using HQ-Net and the ResNet are able to provide important improvements. The CFDG provides a rather marginal improvement. When considering failed test data, it seems that the gesture information remains observable by the model despite the generation errors and smearing. Hence, the ill-trained model tested on a failed test set is able to provide a sufficient recognition rate. Conclusively, the GVIT model, when implemented within the complete proposed methodology, attains a commendable 95.8\% success rate. 
\begin{table}
\centering
\caption{Ablation study of DUR variations for training GViT model}
\label{tb:Gvit}
\begin{tabular}{lc}\toprule
        Models                  & Success rate (\%)  \\\midrule
        DUR w/o \hq             & 91.2 $\pm$ 5.71   \\
        DUR w/o CFDG            & 94.6 $\pm$ 3.32   \\
        DUR w/o ResNet          & 90.03 $\pm$ 4.53  \\
        Failed data             & 89.15 $\pm$ 2.24  \\
        \rowcolor{lightgray}
        DUR                     & 95.8 $\pm$ 4.01   \\
\bottomrule
\end{tabular}
\end{table}


\subsection{Robot demonstration}

\begin{figure}
    \centering
    \includegraphics[width=\linewidth]{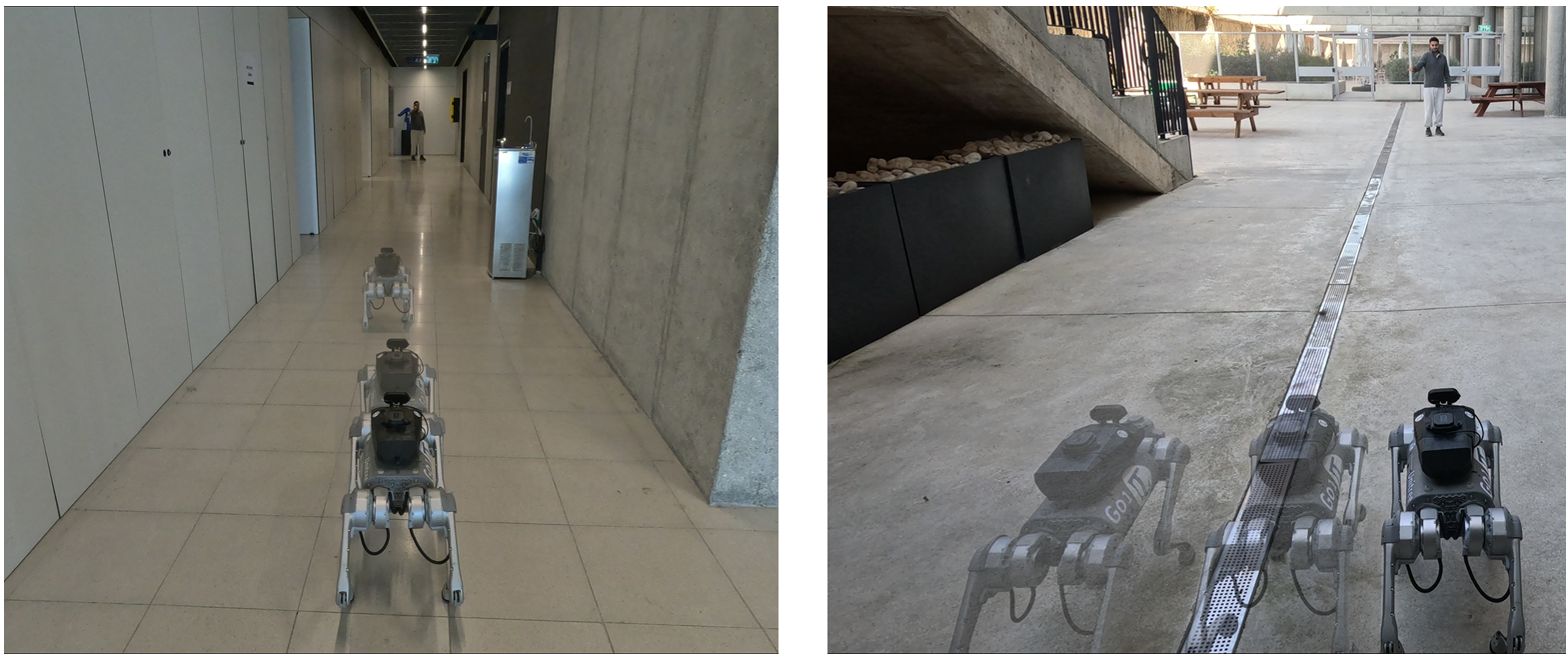} 
    \vspace{-0.6cm}
    \caption{Human directive to a robot through gestures: (left) indoor with beckoning at 23~m, and (right) outdoor with a pointing gesture at 16~m.}
    \label{fig:demo}
    \vspace{-0.5cm}
\end{figure}

The effectiveness and feasibility of the GViT model, trained with only synthetic images, are experimented in directing a Unitree Go1 quadruped robot with gestures. We evaluate the real-time responses of a robot, equipped with a simple RGB camera, to recognize gestures exhibited by a user positioned in ultra-range. For demonstration purposes, a pointing gesture directs the robot to walk sideways; a thumbs-up commands the robot to pitch tilt; a thumbs-down makes the robot lie down; beckoning instructs the robot to walk toward the user; a stop gesture will make the robot stop a previous directive; and with null, the robot will not change its current actions. The robot experiments were conducted in various indoor and outdoor environments as demonstrated in Figure \ref{fig:demo}. A total of 32 gestures were exhibited to the robot with a successful action response rate of 94\%. The results show that the URGR model trained only with data generated by DUR is capable of efficient real-world performance. 
\vspace{-0.3cm}

\label{sec:Evaluation}


\section{Conclusions}

In this paper, we have introduced a comprehensive framework for generating synthetic data of objects in ultra-range distances, addressing the challenges posed by diverse environments, varying distances, and the acquisition of reliable data. The DUR framework seamlessly integrates the diffusion model with the U-Net architecture featuring time embedding to generate high-quality synthetic data. Through extensive experimentation and analysis, we have demonstrated the efficacy of DUR in training a gesture recognition model across different distances, environments and users. By leveraging synthetic data generation, we overcome limitations associated with data acquisition, providing a scalable solution for training gesture recognition models. The major benefit of using DUR is, therefore, the significant reduction of the requirements for real-world data, while capable of training a highly robust recognition model. Future work may include the enforcement of background conditions, enabling the augmentation of an object recognition model with more complex environments. Moreover, exploring the integration of video data and temporal information into the DUR framework holds promise for capturing dynamic changes over time, enhancing the realism and temporal fidelity of synthetic data for recognition models.




\bibliographystyle{IEEEtran}
\bibliography{ref}

\end{document}